# A Neuromorphic Approach To Image Processing And Machine Vision

Arvind Subramaniam
Birla Institute of Technology & Science (BITS) Pilani, Hyderabad Campus
Hyderabad, India
f20150379@hyderabad.bits-pilani.ac.in

*Abstract*— Neuromorphic engineering is essentially the development of artificial systems, such as electronic analog circuits that employ information representations found in biological nervous systems. Despite being faster and more accurate than the human brain, computers lag behind in recognition capability. However, it is envisioned that the advancement in neuromorphics, pertaining to the fields of computer vision and image processing will provide a considerable improvement in the way computers can interpret and analyze information. In this paper, we explore the implementation of visual tasks such as image segmentation, visual attention and object recognition. Moreover, the concept of anisotropic diffusion has been examined followed by a novel approach employing memristors to execute image segmentation. Additionally, we have discussed the role of neuromorphic vision sensors in artificial visual systems and the protocol involved in order to enable asynchronous transmission of signals. Moreover, two widely accepted algorithms that are used to emulate the process of object recognition and visual attention have also been discussed. Throughout the span of this paper, we have emphasized on the employment of non-volatile memory devices such as memristors to realize artificial visual systems. Finally, we discuss about hardware accelerators and wish to represent a case in point for arguing that progress in computer vision may benefit directly from progress in non-volatile memory technology.

Keywords—*Object recognition, Memristor, Image segmentation, Neuromorphic vision sensors, Address Event Representation, Random walker algorithm, Anisotropic Diffusion, Dynamic Vision Sensor.*

## I. INTRODUCTION

The field of neuromorphics is a relatively young one considering the amount of active research that has been invested into it. Despite decades of efforts in the field of artificial vision, no computer algorithm has been able to match the performance and robustness of the primate visual system. Moreover, the proposed algorithms tend to be computationally intensive, and their hardware implementations have encountered problems such as high power consumption and unreliability. It has been discussed that neuromorphic systems offer considerable advantages over conventional computing systems in several areas and introduce the biologically realistic possibility of implementing collective computation and memory storage simultaneously. Autonomous visual and auditory processing systems can benefit greatly from the low-power features of the neuromorphic hardware technology.

An important processing stage in many visual processing systems is image segmentation. It is the process of dividing an image into multiple parts, thereby making it easier to identify the foreground or relevant information present in the image. Previous studies have proposed the use of neuromorphic circuits [1], which can perform object-based selection and segmentation. Recent studies have applied the concept of synchronized oscillations, where photo-detector cells receiving similar light intensities from an object oscillate together in a synchronous manner [2], [3]. Visual tasks such as object-based selection and segmentation have been implemented with the help of specialized sensory processing functions called Neuromorphic Vision sensors. Several publications have appeared in recent years documenting the use of neuromorphic vision sensors and pre-processors to integrate imaging, color segmentation and color-based object recognition [4]–[6]. These sensors consist of analogue electronic circuits operating in the sub-threshold regime, interfaced to digital processing systems that execute machine vision algorithms in order to realize selective attention, object recognition, etc. in artificial systems. Recent developments in the field of neuroscience have led to a renewed interest in the development of computational models and algorithms for auditory and visual perceptions in the human brain. Among these models, the Saliency based model and the HMAX model have been extensively used for visual attention and object recognition respectively. The aim of this paper is to provide a comprehensive exploration of some of the recent techniques in the field of neuromorphics, which have been employed to implement visual tasks such as selective attention, gesture recognition, object recognition and tracking.

The remainder of the paper is organized into the following sections. Section II outlines a few techniques and approaches adopted to execute scene segmentation. In Section III, we examine the types and mechanisms of neuromorphic sensors. Section IV describes the models and algorithms used to better implement the visual challenges such as object recognition and selective attention in artificial systems. Furthermore, a modest introduction to hardware accelerators has also been presented. Finally, Section V draws conclusions.

## II. IMAGE SEGMENTATION

There are several popular image segmentation algorithms based on a variety of techniques such as thresholding, edge detection, clustering, partial-differential equations (PDEs) and artificial neural networks [7], [8].These methods can be categorized into supervised, semi-supervised and unsupervised. The Random walker algorithm is a semi-supervised algorithm that models the image as a graph where pixels correspond to nodes and are connected to neighboring pixels through weighted edges. Recent studies have also employed unsupervised random walk approaches to execute image segmentation. The edges are weighted in accordance with the similarity between the corresponding pixels, and the edge weights are equal to electrical conductance.

In the first- among other interpretations of the algorithm-the subject labels a small number of pixels commonly typified as seeds. Each unlabelled pixel is assumed to release a random walker, after which the probability that a random walker will first arrive at each seed is calculated. After the probability for every pixel is calculated, each pixel is assigned to the respective seed which has the maximum probability of receiving a random walker from the corresponding pixel. Segmentation of the image is complete when all the unlabelled nodes have each been assigned a seed. The random walker has been shown to achieve segmentation in multiple ways such as solving a discrete Dirichlet problem [9] and employing an iterative method to solve a discrete PDE [10]. The former approach, however, has proven to be computationally intensive for real time applications. In an alternative approach, the nodes corresponding to each pixel are assigned to a seed based on the effective conductance (edge weights) between the nodes and the foreground or background seeds. After a few iterations, pixels with intensities higher than a certain threshold value are assigned to foreground seeds, else treated as background. A method which is slightly different from the aforementioned approaches involves viewing the graph of the image as an electrical circuit and establishing an electrical potential at each node associated with the foreground. The magnitude of these potentials would be equal to the probability that a random walker dropped at a node will reach a foreground seed before reaching a background seed. However, in case of iterative algorithms, the conductance between nodes/pixels must change appropriately after every iteration. This is not practically realizable in conventional CMOS, since one would have to manually change the resistance offered by the device connecting adjacent nodes/pixels.

Yet another interesting approach to the Random walker algorithm involves the concept of anisotropic diffusion [11]. Anisotropic diffusion is a method of denoising images by smoothing the pixel values only on one side of the boundary, as opposed to Isotropic diffusion, which averages pixel values across edges. The latter technique is applied in local averaging filters and median filters. The main drawback of the isotropic diffusion is the loss of weak boundaries and the undesirable smoothing of all edges. There have also been modifications to anisotropic diffusion such as the introduction of an edge seeking diffusion coefficient (varies with image position) that prohibits diffusion across strong edges [12].

A novel approach would be to employ memristor-based crossbar arrays to represent edges between a node and a foreground/background seed, as a memristor can change its effective conductance based on the history of current that had previously flown through it. As a result, the variable resistance of the memristor would allow for changes in the edge weights when there is a change in the potential difference across nodes after every iteration. Hence, this would eliminate the need to manually change the resistance of the device connected across two nodes or pixels.

III. NEUROMORPHIC OSCILLATOR NETWORKS

Previous research has documented the use of oscillatory correlation to segment and represent an image [13], [14]. In such a scheme, different objects present in the image are represented by different groups of synchronized oscillators. As a result, no two distinct objects correspond to the same set of synchronized oscillations. Furthermore, the locally excitatory globally inhibitory oscillator network (LEGION) has been shown to provide a viable and effective framework to solve the problem of image segmentation [15], [16]. A major drawback of the LEGION algorithms employed in previous research has been their sensitivity to noise. Executing the algorithm without any modifications would lead to the problem of fragmentation since noisy images may result in a large number of fragments. However, this problem can be solved by suppressing the oscillators corresponding to the noisy regions of the image.

Previous studies have indicated that memristors have proven to be useful in designing non-linear oscillators [17]–[19] and contribute to the synchronization of coupled neuromorphic systems to perform visual segmentation [20]–[22]. However, there are a few challenges to the application of memristors in designing oscillators. The resistance offered by the memristor is a non-linear function of time. However, this might be resolved by using a memristor-transistor pair in the circuit.

IV. NEUROMORPHIC VISION SENSORS

Neuromorphic vision sensors have been in the forefront of the development of artificial visual systems, as they offer visual perception at a lower computational load. Several studies have discussed the emulation of the human retina with the help of analog electronic circuits [23]–[25].These circuits are parallel and operate in the sub-threshold domain. As a result, two or more computational problems can be solved simultaneously, while consuming extremely low amounts of power. This permits the use of circuits with high computational density. Furthermore, the circuits are asynchronous i.e. the transition from one state to another is the result of a change in their primary inputs. Despite issues regarding instability and inefficiency, they are preferred over synchronous circuits due to the continuous nature of the sensory input. Moreover, the employment of synchronous systems would lead to misinterpretation of sensory information as a result of aliasing.

The concept of neuromorphic sensors in not new. Since the 1990s, several studies have explored and widely proposed two types of neuromorphic sensors, namely Silicon retinas [26], [27] and silicon cochleae [28], [29]. These systems make use of a protocol called Address Event Representation (AER) to transmit signals asynchronously. In AER, a variable number of lines (bus) are used to transmit data. This data is usually an address that has been assigned to each analog element present on the sending device [30]. The ACK and REQ lines are active low lines which enable the synchronization of data between the sender and the receiver. A major advantage of using AER is that the power consumed for the transmission of sensory information is significantly reduced, since AER sensors transmit signals based on the activity of each individual pixel, contrary to CMOS sensors. The activity of a pixel is represented by a stream of digital pulses produced by the neuromorphic sensor.

Asynchronous devices that respond only to a change in the brightness of a pixel, such as the Dynamic Vision Sensor (DVS) have been proposed, their key advantages being a significant reduction in data storage and computational complexity. Several studies have employed Spike-timing-dependent plasticity (STDP) to extract correlated features (temporally) from dynamic vision sensors [31], [32]. STDP is a biologically-realistic learning mechanism based on the relative timing of the post- and pre-synaptic spikes.

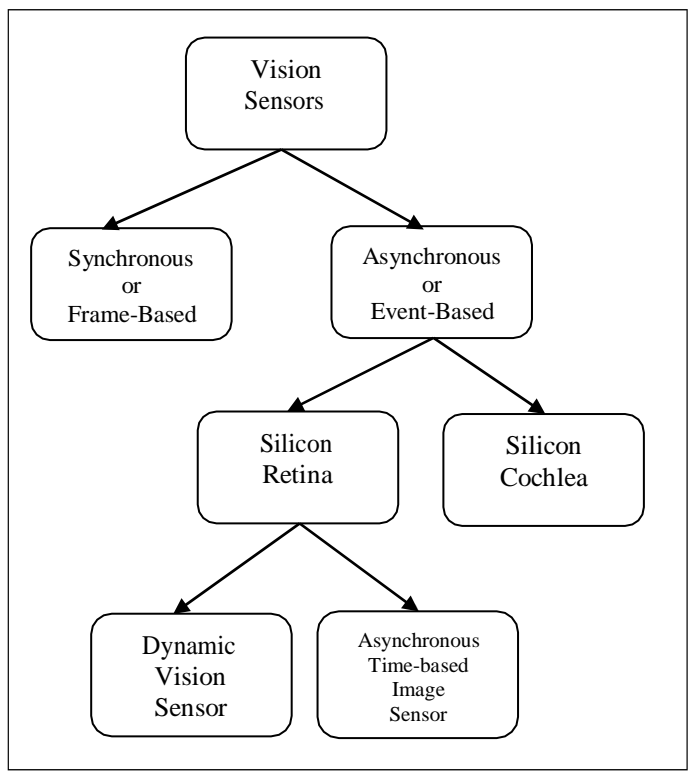

**Figure 1. Types of vision sensors**

Several studies have emphasized on the utilization of memristors as a modulating synapse between neurons, in order to realize Spike-timing-dependent plasticity in neuromorphic systems [33]–[37], since the conductance of a memristor can be modified by controlling the charge through it. The most recent silicon retina prototype is the Asynchronous time-based Image Sensor (ATIS) which has the feature of conditional exposure measurement in addition to measuring a change in brightness.

## V. NEUROMORPHIC VISION ALGORITHMS FOR ATTENTION AND OBJECT RECOGNITION

### A. Visual Attention

The mammalian visual system has an inherent ability to detect the salient and important sections of an image and filter out redundant visual details through the selective attention mechanism. A number of models and algorithms of the human visual system have been proposed [38]. The focus of these systems is to direct attention to regions of interest in an image. As a result, further processing to sub-regions of the image is restricted and the amount of data for complex processing tasks such as object recognition is reduced. The locations to be analyzed are selected with the help of two guiding influences: the goal-oriented, top-down attention and the image-driven, bottom-up attention [39].

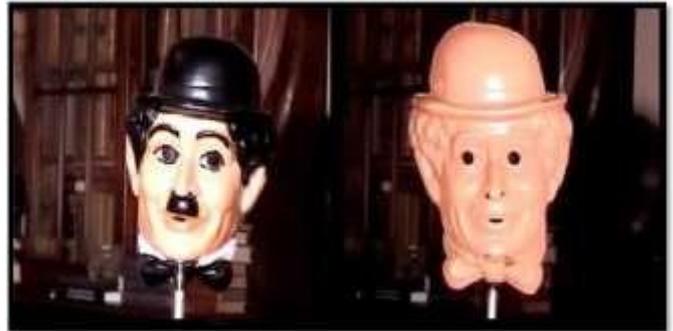

**Figure 2**

Top-down attention (TD), also known as sustained or endogenous attention involves the voluntary allocation of attention to certain features or regions in space for sustained periods of time. As a result, it is slower and requires more effort to engage than the bottom-up attention, in which the resources are involuntarily captured [40], [41]. Top-down or sustained attention would be utilized when the subject intentionally directs attention to a task such as driving, etc. Alternatively, an unexpected event such as an accident in the adjacent lane, might inadvertently capture the subject's attention, as it is significantly different from the surrounding events. This involuntary allocation of attention is influenced by the bottom-up attention (BU). Recent studies have documented two types of top-down attention (TD) mechanisms, namely the volitional top-down selection process and the mandatory top-down selection process [42]. The former mechanism is an attentional TD process that exerts influence on sensory processing through acts of will, such as wilfully shifting attention a particular region of space (top-down spatial attention) in the middle or to all black items (top-down feature attention) [43], [44].

Alternatively, the mandatory TD process influences sensory processing in an automatic and persistent manner. Fig.2 presents the mental impression of the stimulus, which can be wilfully changed from a scene to a face and vice-versa, demonstrating TD modulation of sensory processing in a dynamic manner.

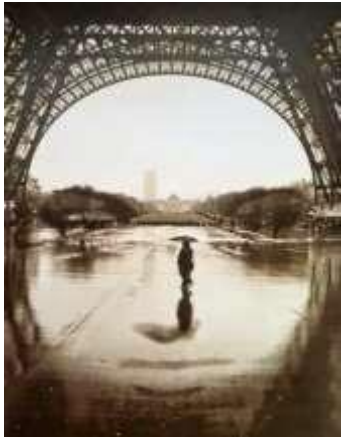

**Figure 3**

On the other hand, in Fig.3, which presents the convex and concave sides of a mask, the subject would be unable to distinguish the concave side (inside) of the mask as it would still seem to be convex. This inherent bias in identifying faces as convex rather than concave is an illuminative example of mandatory TD processing.

The Bottom-up attention (BU), also typified as transient or exogenous attention, is instrumental in the realization of involuntary attention in artificial visual systems. Central to the discipline of the bottom-up attention has been the saliency-based model of visual attention [45]. The model employs feature maps to detect regions of interest in an image based on spatial discontinuities in features such as orientation, color, intensity, etc. and combines them into a unique topographical saliency map. The saliency map is then scanned sequentially in order of decreasing saliency. This breaks down the complexity of the image and enables real-time performance. Therefore, it be rapidly implemented as opposed to top-down attention, which requires successive operations on sensory input [46].

However, despite the real-time performance demonstrated by the bottom-up influence, there is a need to combine both the influences in order to successfully execute tasks such as visual surveillance, where the detection of goal-specific targets as well as unexpected visual events is critical.

Certain studies have revealed the integration of both the afore-mentioned guiding influences to optimize detection speed [47], [48]. First, the bottom-up influence determines the visual salience of each object in a number of dissimilar feature maps. Next, the top-down guiding influence makes use of learnt knowledge of the aforementioned features of the object in order to enhance the relative weights of the bottom-up maps. This maximizes the overall salience of the visual target with respect to the surrounding. Therefore, the target is more salient than other visual distractions and target detection speed is maximized. The classification of the afore-mentioned neuromorphic vision approaches has been illustrated in Fig.4. 'Both' refers to the integration of the Top-down and Bottom-up guiding influences.

### B. Object Recognition

Visual Object recognition is the ability of the mammalian brain to perceive and identify an object's physical properties. Therefore, it is a crucial feature of any artificial visual processing system. Despite several attempts to practically implement object recognition, it has proven to be an extremely complex computational problem.

The computational load lies in the fact that each object can project infinitely many dissimilar 2-D images as the size, orientation and background of the object vary with respect to the subject [49]. Among the various models proposed, the Hierarchical Model and X (HMAX) model has been extensively used for image classification with applications in object and gesture recognition [50], [51].

The HMAX model is a four-level feed forward architecture fundamentally made up of the following Simple (S) and Complex (C) layers, which perform convolution and max-pooling operations to build complex features:

The first layer, commonly known as the S1 layer, is the result of processing 12 two-dimensional Gabor filters of dimensions 11x11, with different orientations at every possible pixel. Next, an image pyramid corresponding to each Gabor filter is created. Each pyramid consists of a set of 12 scales, where each scale is smaller than the previous scale by a factor of $2^{1/4}$. The 12 image pyramids (144 scales) are then fed into the second layer.

The second layer, referred to as the C1 layer, convolves the image pyramids having the same orientation with a Max filter, to create scale and position invariance over local regions i.e., the system is unchanged to changes in scale and position. Further, the C1 layer also subsamples the pyramids by a factor of 5. As a result, C1 pyramids are smaller despite having the same number of orientations.

The third layer, commonly typified as the S2 layer, is one of the most important and computationally intensive layers and accounts for over 80% of the total execution time. This arises as a result of convolving every scale of each multi-scaled pyramid from the C1 layer. In other words, the S1 layer processes the results of the C1 layer by matching it against a set of pre-learned feature prototypes. These feature prototypes are considerably more detailed than the previously mentioned features detected by the S1 layer.

Finally, the C2 layer removes all scale and position information by pooling the obtained results across different scales and orientations to produce feature vectors.

Simply put, the objective of the aforementioned model is to obtain feature vectors from gray scale images, and classify those vectors with the help of a Support Vector Machine (SVM).

However, there have been a number of problems accompanying the implementation of the HMAX model such as low energy efficiency and excessive power consumption. Recent research has documented the use of hardware accelerators, which are in the form of programmable cores or programmable logic such as FPGA, ASIC and GPU, to primarily speed up the S2/C2 stage as it is the most time-consuming stage. It has been indicated that FPGA provides the best functional configurability while ASIC demonstrates the highest efficiency.

Recent studies have employed memristive Neuromorphic Computing Accelerators (NCA) consisting of memristor-based crossbar arrays, as opposed to conventional accelerators which are based on systolic arrays. Memristor-based models of the Boltzmann Machine, a massively parallel computational model used for solving combinatorial optimization problems, have been proposed to accelerate neural computation tasks in an energy efficient manner. The memristive hardware accelerator had an appreciable improvement of 6.89x and 5.2x on performance and power consumption respectively, as compared to a standard RRAM based memory [52]. However, the proposed memristor-based accelerator has not been shown to solve problems of higher computational complexity yet.

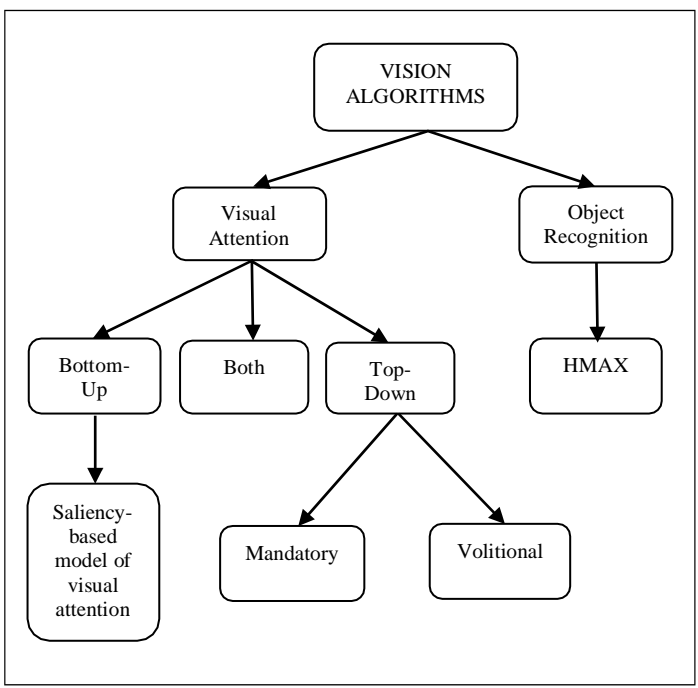

Figure 4. Classification of Neuromorphic vision algorithms

VI. CONCLUSION

This paper focuses on the techniques and algorithms employed to practically implement critical visual tasks such as image segmentation, selective attention and object recognition, by making use techniques pertaining to the emerging field of neuromorphics. We have also recommended a novel approach which involves utilizing memristor-based crossbar arrays to act as edges between a node and a foreground/background seed and hope to instigate new direction for further research integrating neuromorphics, image processing and non-volatile memory devices. Apart from the realization of the aforementioned tasks, the mechanisms and types of neuromorphic vision sensors have also been explored. We have laid emphasis on the use of non-volatile memory devices such as memristors to enhance performance and energy efficiency during the execution of the aforementioned tasks. The application of memristors in the implementation of image segmentation, neuromorphic sensors and hardware accelerators has proven to be more energy efficient than previous methods, which used conventional CMOS.

Throughout the span of the paper, we have tried to highlight the importance of using memristors to better solve challenges pertaining to image processing and computer vision.

In addition to the employment of memristors to overcome visual challenges, future research directions should focus on the application of memristive systems to solve issues related to upcoming fields such as the Internet of Things (IoT) and quantum computing.